%% file: main.tex
\documentclass[10pt,twocolumn,letterpaper]{article}
\usepackage[pagenumbers]{cvpr} 
\usepackage{array}
\usepackage{booktabs}
\usepackage{graphicx}
\usepackage{makecell}
\usepackage{amsmath}
\usepackage{bm}
\input{preamble}

\newcommand{\beforefigcaption}{\vspace{-8mm}}
\newcommand{\afterfigcaption}{\vspace{-4mm}}
\newcommand{\beforetab}{\vspace{-3mm}}
\newcommand{\aftertab}{\vspace{-3mm}}
\newcommand{\beforesection}{\vspace{-2mm}} 
\newcommand{\aftersection}{\vspace{-2mm}}
\newcommand{\beforesubsection}{\vspace{-1.5mm}}
\newcommand{\aftersubsection}{\vspace{-1.5mm}}

\definecolor{cvprblue}{rgb}{0.21,0.49,0.74}
\usepackage[pagebackref,breaklinks,colorlinks,citecolor=cvprblue]{hyperref}
\def\paperID{6112} 
\def\confName{CVPR}
\def\confYear{2024}

\title{Tri-Modal Motion Retrieval by Learning a Joint Embedding Space}

\author{
  Kangning Yin$^{1}$ \quad
  Shihao Zou$^{2}$ \quad
  Yuxuan Ge$^{1}$ \quad
  Zheng Tian$^{1}$\thanks{$^{}$Correspondence to Zheng Tian \quad \href{mailto:tianzheng@shanghaitech.edu.cn}{\color{black}$<$tianzheng@shanghaitech.edu.cn$>$}} \\
  $^{1}$ ShanghaiTech University \quad
  $^{2}$ Shenzhen Institute of Advanced Technology, Chinese Academy of Sciences\\
  {\tt\small \{yinkn2022, geyx2023, tianzheng\}@shanghaitech.edu.cn}, \quad
  {\tt\small {sh.zou@siat.ac.cn}}
}
\begin{document}
\maketitle
\input{sec/abstract_new}
\input{sec/intro_new}
\input{sec/2_relatedwork}
\input{sec/method_new}
\input{sec/4_experiment}
\input{sec/5_conclution_discussion}

{
    \small
    \bibliographystyle{ieeenat_fullname}
    \bibliography{main}
}


\end{document}

%% file: preamble.tex
\usepackage[dvipsnames]{xcolor}


%% file: sec/abstract_new.tex
\begin{abstract}
Information retrieval is an ever-evolving and crucial research domain. The substantial demand for high-quality human motion data especially in online acquirement has led to a surge in human motion research works. Prior works have mainly concentrated on dual-modality learning, such as text and motion tasks, but three-modality learning has been rarely explored. Intuitively, an extra introduced modality can enrich a model’s application scenario, and more importantly, an adequate choice of the extra modality can also act as an intermediary and enhance the alignment between the other two disparate modalities. In this work, we introduce LAVIMO (LAnguage-VIdeo-MOtion alignment), a novel framework for three-modality learning integrating human-centric videos as an additional modality, thereby effectively bridging the gap between text and motion. Moreover, our approach leverages a specially designed attention mechanism to foster enhanced alignment and synergistic effects among text, video, and motion modalities. Empirically, our results on the HumanML3D and KIT-ML datasets show that LAVIMO achieves state-of-the-art performance in various motion-related cross-modal retrieval tasks, including text-to-motion, motion-to-text, video-to-motion and motion-to-video.
\end{abstract}
\label{abstract}

%% file: sec/intro_new.tex
\beforesection
\section{Introduction}
\aftersection


Recent technologies such as Generative Adversarial Networks (GANs)~\cite{goodfellow2014generative}, diffusion models~\cite{ho2020denoising, ho2022classifier, song2020denoising} and multi-modal models~\cite{ramesh2021zero, wang2023largescale} have achieved significant advancements. This progress has spurred a surge in human motion related research including motion generation~\cite{tevet2022human, yuan2023physdiff, wang2023fg} and motion retrieval~\cite{petrovich23tmr, kapadia2013efficient, wang2016adaptive}. Motion generation, while effective in generating realistic motion conditioned on textual data~\cite{guo2022tm2t, zhang2022motiondiffuse, tevet2022human, petrovich2022temos}, music~\cite{zhuang2022music2dance, tseng2023edge} and motion sequences~\cite{yan2019convolutional, zhao2020bayesian, sasagawa2021motion, raab2023modi}, it often struggles with generating diverse or contextually appropriate motions in complex scenarios and offers little controllability~\cite{zhu2023human}. 
In contrast, motion retrieval completes motion acquirement through retrieving specific human motion sequences from a large dataset or database given queries commonly in forms of text, motion sequences, text, images, etc. It addresses the above challenges by leveraging external databases with high quality and diversity. Therefore, it is particularly valuable in industries with stringent requirements for precision, realism and controllability such as character animation~\cite{holden2017phase, bergamin2019drecon} , virtual fitting rooms~\cite{han2018viton} and film production~\cite{muhling2017deep}. 

\begin{figure}
    \centering
    \includegraphics[width=0.5\textwidth]{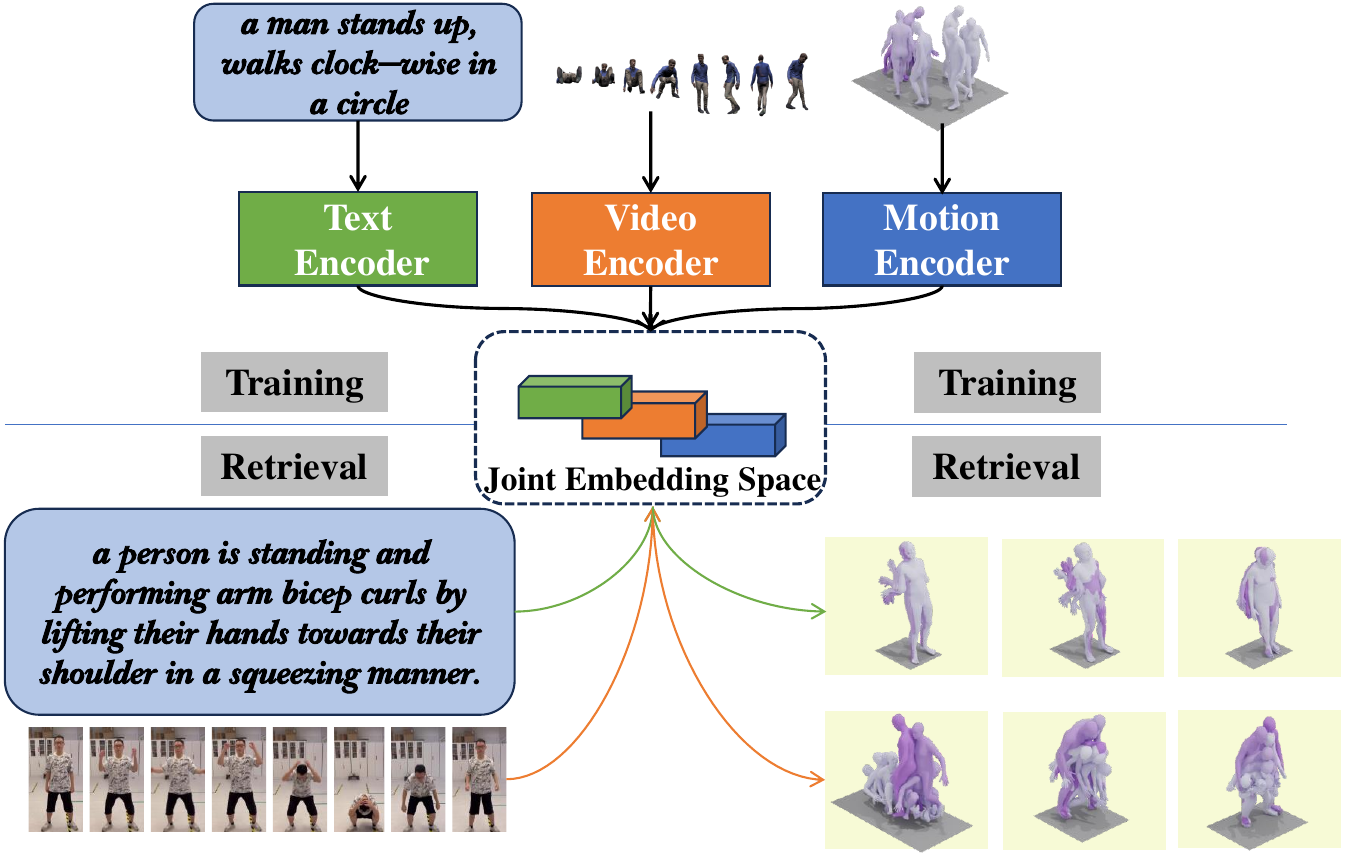}
    \beforefigcaption
    \caption{\textbf{Overview of LAVIMO.} During the training phase, the three modalities are processed through their distinct encoders. Subsequently, the resultant embeddings are aligned within a unified joint embedding space utilizing contrastive learning techniques. In the inference stage, the model is capable of accepting texts or videos as input queries, enabling the retrieval of corresponding motion data effectively.}
    \afterfigcaption
    \label{fig:1}
\end{figure}


Many effective prior works in motion retrieval focus on dual-modality problems, especially text and motion related tasks given the generality of language. For example, TMR~\cite{petrovich23tmr} utilizes contrastive learning to construct a joint embedding space for text-to-motion and motion-to-text retrieval. However, prior works often face a challenge that the spatial distance between two modalities such as text and motion being considerably vast~\cite{zhang2023t2m}, thus merging them into a unified embedding space often requires large datasets with high quality labels, which are also scarce. In comparison, the human-centric video modality serve as a compact, low-dimensional representation of intricate 3D motions~\cite{zheng2023deep}, placing it closer in the spatial spectrum to motion. On the other hand, the availability of numerous video-text datasets bolsters the integration of the human-centric video and text modalities. Therefore, in this work, we take video as an intermediary modality to effectively narrow the spatial distance between each pair of the three modalities. Similar ideas can also be seen in other domain, for example, VALOR~\cite{chen2023valor} leverages audio as an auxiliary modality to bolster the alignment between video and language. Building upon the integration of video into a cross-modal learning framework, we can also empower the model to execute video-to-motion retrieval task.

Conventional multi-modal learning frameworks often include a reconstruction branch to ensure that the information is not lost when translating from one modality to another. However, these frameworks tend to concentrate on data from only one specific modality, overlooking the distinct information inherent in other modalities. For instance, text contains specific details such as intentions and action paces, which are absent in video and motion modalities. Similarly, video offers unique visual information that is absent in other formats. In this work, we introduce an attention mechanism in which motion is utilized as queries. This mechanism is designed to extract relevant information from texts and videos, thereby addressing the limitations mentioned above. To the best of our knowledge, this paper first proposes to leverage video as an additional modality to enhance the alignment between the three modalities.

Our key contributions are summarized as follows: (i) We introduce LAnguage-VIdeo-MOtion Alignment (LAVIMO), a framework designed to cultivate a cohesive embedding space across the three aforementioned modalities. This architecture can accomplish both text-to-motion retrieval and video-to-motion retrieval tasks. 
(ii) We create a custom attention mechanism to merge the three modalities in our motion reconstruction branch, enabling us to incorporate valuable information from text and video when reconstructing motion sequences. This approach is intended to improve the alignment between the three modalities.
(iii) In our efforts to enrich the available data, we augment both the HumanML3D and KIT datasets with RGB videos. This is achieved by animating and rendering avatars corresponding to specific motions. Such augmentation paves the way for more expansive and informed multi-modal research in human motion dynamics.

%% file: sec/2_relatedwork.tex
\beforesection
\section{Related Work}
\aftersection
\label{sec:relatedwork}
\noindent{\bf Motion Generation.} Motion generation is a popular topic in recent years. Existing frameworks for motion generation can be broadly categorized into two types. The first is a joint representation framework, in which text inputs are first processed by a pre-trained text encoder. Subsequently, these processed inputs are passed through a specially designed architecture, such as a diffusion model, along with the original motion sequences. This process aims to effectively reconstruct motion sequences. The representative works in this category are MDM~\cite{tevet2022human} and T2M-GPT~\cite{zhang2023t2m}. The second so-called coordinated representation framework focuses on constructing a joint embedding space between texts and motions leveraging  Auto-Encoders (AE) and Variational Auto-Encoders (VAE). The representative works in this category are MotionCLIP~\cite{tevet2022motionclip}, TM2T~\cite{guo2022tm2t} and TEMOS~\cite{petrovich2022temos}. Our work is different from them as we incorporate a third modality and align the three modalities in a joint space.


\noindent\textbf{Motion retrieval.} Motion retrieval has gained popularity due to the rapid growth of motion capture data~\cite{fern2012biomechanical, wei2012accurate, mahmood2019amass}. It offers an alternative method for obtaining motion sequences. 
Its primary goal is to identify and extract the motion sequence that exhibits the highest degree of similarity from an extensive motion database. 
Previous works such as DreCon~\cite{bergamin2019drecon} and PFNN~\cite{holden2017phase} mainly focus on motion-to-motion retrieval, the specific motion is retrieved according to movement direction, heading direction, speed and locomotion style. 
However, motions are often represented parametrically, which can be challenging for individuals who are not experts in motion related research to comprehend.
To address this limitation, TMR~\cite{petrovich23tmr} initiates a novel branch termed as text-to-motion retrieval. This approach involves creating a cross-modal embedding space using contrastive learning. Additionally, TMR employs a negative filtering technique, ensuring that texts with similar meanings are considered as positive pairs. Nicola et al.~\cite{messina2023text} have also explored this field in their work. The primary goal of text-to-motion retrieval is to search for motion sequences in a database based on a specific text query. The performance of TMR has not reached its potential compared to that in the image and language domain. This is largely due to the lack of motion datasets. We incorporate an additional video modality to bridge the gap between text and motion, thereby reducing the reliance on extensive datasets.
\noindent\textbf{Multi-Modal Learning.} In the initial stages of multi-modal research, two common techniques are early fusion and late fusion~\cite{snoek2005early, gunes2005affect, gadzicki2020early}. 
In early fusion, features from different modalities are combined before being fed to a classifier. 
While in late fusion, features are processed separately, and their outputs are combined only in the final stages.
In recent years, the field of multi-modal learning has experienced rapid advancements. 
Specifically, frameworks such as CLIP~\cite{DBLP:journals/corr/abs-2103-00020}, DALL-E~\cite{DBLP:journals/corr/abs-2102-12092} and BERT~\cite{DBLP:journals/corr/abs-1810-04805} concentrate on elucidating the relationship between text and image. 
MV-GPT~\cite{seo2022end}, Cap4Video~\cite{wu2023cap4video} and EMCL~\cite{jin2022expectation} adeptly bridge the connection between text and video. 
Most of the works mentioned above utilize contrastive learning to build a joint embedding space between two modalities. 
Recent advancements in the video-language understanding domain start to leverage auxiliary modalities to enhance the alignment between video and text. Among these, Yusuf et al.~\cite{DBLP:journals/corr/AytarVT17} effectively utilizes a teacher-student model and ranking loss to develop a cross-modal representation bridging text, image, and audio. Mithun et al.~\cite{mithun2018learning} skillfully integrates image appearance features, temporal dynamics, and audio cues from videos to enrich the information pool. VALOR~\cite{chen2023valor}, which introduces a vision-audio-language pre-training model stands out as a notable example. 

%% file: sec/method_new.tex
\beforesection
\section{Method}
\aftersection
Our primary goal is to train a joint embedding space of three modalities, \ie, text, video and motion, ideally allowing them to mutually represent each other. With this embedding space, we can accomplish various tasks involving text-to-motion retrieval or video-to-motion retrieval. In this section, we begin by providing a comprehensive definition of the three related modalities along with their distinct encoders in Sec.~\ref{sec3.1}. Following that, we present the three-modality co-training process in Sec.~\ref{sec3.2} and the description of motion-related retrieval tasks in Sec.~\ref{sec3.3}.

\beforesubsection
\subsection{Modality Definitions and Model Architecture}
\aftersubsection
\label{sec3.1}
Our extensive model architecture includes a motion encoder, a text encoder, and a video encoder. Each encoder independently extracts features from motion, text, and video, which are then aligned in a joint embedding space. To accelerate convergence, we initialized these encoders with pre-trained models.

\noindent{\bf Motion Encoder.} The motion sequence comprises a series of 3D human poses $P=[P_1, P_2, ..., P_{l_m}]$, where $l_m$ represents the temporal length of a motion sequence and $P \in {\mathbb{R}}^{l_m \times c_m}$ with $c_m$ being the feature dimension of a single pose. Each pose is parameterized using the SMPL model~\cite{SMPL:2015}. We follow the motion representation used in~\cite{guo2022generating}, which consists of joint velocities, local joint positions, 6D form local joint rotations and the foot contact label. We utilize the motion encoder from MotionCLIP~\cite{tevet2022motionclip} in our work. 
In this process, the sequence $P$ undergoes a linear projection to map into the latent space. A [CLS] token, representing the global feature of the motion sequence, is then appended at the beginning. This processed sequence is fed into the transformer encoder. We use the first position of the output as the representative motion feature, denoted as $\bm{{e}_m}\in \mathbb{R}^{B\times C}$, where $B$ is batch size and $C$ is the latent dimension.

\noindent{\bf Text Encoder.} 
The text description in our model is a sequence of words $T=[W_1, W_2, ..., W_{l_t}]$ that succinctly summarize the contents of the motion, where $l_t$ is the number of words. This description typically includes elements such as the target character (\eg,``The man''), specific actions performed by the character (\eg, ``doing jumping jacks''), and stylistic attributes that characterize these actions (\eg, ``at a fast speed'', ``counterclockwise''). Additionally, a single text description can involve multiple actions arranged in a sequence (\eg, "The man walks forward at a rapid pace and then makes a right turn.").
We employ DistilBERT~\cite{DBLP:journals/corr/abs-1910-01108} as the foundational model for text processing. Textual tokens are generated through the DistilBERT tokenizer. Following the architecture of the motion encoder, a [CLS] token is appended at the beginning of the token sequence. This entire sequence is then processed through the transformer encoder. The textual feature representation is extracted from the first output position, denoted as $\bm{{e}_t}\in \mathbb{R}^{B\times C}$.


\noindent{\bf Video Encoder.} A video is represented as a sequence of frames $I=[I_1, I_2, ..., I_{l_v}]$, where we uniformly sample $l_v$ frames from an RGB video as the input to avoid redundancy. The video encoder consists of a CLIP~\cite{ramesh2021zero} image encoder and a temporal transformer. 
Initially, the video sequence is processed through the CLIP pre-trained image encoder to extract image features for each selected frame. Subsequently, considering the temporal aspects of the video, we input these image features into a temporal transformer. The video embedding is obtained by applying average pooling to the output of this temporal transformer, denoted as $\bm{{e}_v}\in \mathbb{R}^{B\times C}$. 

\beforesubsection
\subsection{Cross-Modality Contrastive Learning}
\aftersubsection
\label{sec3.2}
\noindent{\bf Multi-Modality Alignment.} We employ contrastive learning to construct a fine-grained embedding space that bridges the three modalities. This alignment is achieved by minimizing the following loss function:
\begin{equation}
    \mathcal{L}_{align} = \mathcal{L}_{align}^{mt} + \mathcal{L}_{align}^{mv} + \mathcal{L}_{align}^{tv},
\end{equation}
where $\mathcal{L}_{align}^{xy}$ is the alignment loss between modality $x$ and modality $y$. 

Taking motion and text as an example, to maximize the proximity between positive samples while minimizing it between negative samples, we use the Kullback–Leibler (KL) divergence loss to establish our joint embedding space between text and motion:
\begin{equation}
\mathcal{L}_{align}^{mt} = KL\left({S_{pred}^{t2m}}, {S_{target}}
\right)+KL\left({S_{pred}^{m2t}}, S^{\top}_{target}\right),
\end{equation}
where $S_{pred}^{t2m}$ and $S_{pred}^{m2t}$ represent the similarity matrix between text to motion and motion to text respectively. These can be calculated as follows: 
\begin{equation}
S_{pred}^{t2m}(i, j) = \frac{\exp(cos(\bm{{e}_{t}}^i, {e}_{m}^i)/ \tau)}{\sum_{j=1}^{B} \exp(cos(\bm{{e}_{t}}^i, e_{m}^j)/ \tau)},
\end{equation}
\begin{equation}
S_{pred}^{m2t}(i, j) = \frac{\exp(cos(\bm{{e}_{m}}^i, {e}_{t}^i)/ \tau)}{\sum_{j=1}^{B} \exp(cos(\bm{{e}_{m}}^i, {e}_{t}^j)/ \tau)},
\end{equation}
where $\tau$ is the learnable temperature parameter, the indices $(i, j)$ refer to the specific element within the respective matrix and the function $cos(\cdot)$ represents the cosine similarity between the two input embeddings:
\begin{equation}
    cos(\bm{e}^{i}, \bm{e}^{j})= \frac{ \bm{e}^{i} \cdot \bm{e}^{j\top} }{ \lVert \bm{e}^{i} \rVert \lVert \bm{e}^{j} \rVert }.
\end{equation}

Meanwhile, $S_{target}$ is the target similarity matrix. 
However, considering that different text descriptions might convey similar meanings, defining $S_{target}$ strictly as an identity matrix can pose issues. Such a definition would cause the model to incorrectly categorize texts with similar meanings as negative pairs, which could hinder the model's efficacy in text-motion retrieval tasks.
To address the aforementioned issue, we adopt the negative filtering technique proposed by TMR~\cite{petrovich23tmr}. 
Specifically, we employ a pre-trained language model~\cite{DBLP:journals/corr/abs-2102-07033} to encode text labels into embeddings, which serve as the ground truth, denoted as $\bm{\hat{e}}_t$. 
Subsequently, we calculate the cosine similarity between pairs of text embeddings within a single batch of size $B$. We define a threshold, $\epsilon$, to represent the minimum similarity score required for a pair to be considered positive. Unlike the method used in TMR, instead of discarding scores above the threshold $\epsilon$, we retain these scores in $S_{target}$. 
Therefore, we establish the target similarity matrix $S_{target}$ with each element defined as:
\begin{equation}
    S_{target}(i,j) = 
    \left\{
        \begin{aligned}
        & cos(\bm{\hat{e}}_t^{i}, \bm{\hat{e}}_t^{j}), & \text{if }  cos(\bm{\hat{e}}_t^{i}, \bm{\hat{e}}_t^{j}) \ge \epsilon, \\
        & 0, & \text{otherwise}.
        \end{aligned}
    \right.
\end{equation}

\noindent{\bf Features Fusion.} 
In this module, we incorporate a specially designed attention mechanism to blend features effectively during the motion reconstruction process. We treat motion embeddings as queries, which actively seek relevant information across textual and visual domains.
The process can be expressed by:
\begin{equation}
    \begin{aligned}
        \bm{\Tilde{e}}_m & = Atten(Q_{\bm{{e}_m}}, K_{\bm{{e}_m}}, V_{\bm{{e}_m}}) + Atten(Q_{\bm{{e}_m}}, K_{\bm{{e}_t}}, V_{\bm{{e}_t}})\\
                    & + Atten(Q_{\bm{{e}_m}}, K_{\bm{{e}_f}}, V_{\bm{{e}_f}}), 
    \end{aligned}
\end{equation}
where $Atten(\cdot)$ is self-attention operations proposed in~\cite{vaswani2017attention}, and $Q$, $K$, $V$ represent query, key and value, respectively. The subscripts $\bm{e}_m$, $\bm{e}_t$ and $\bm{e}_v$ indicate the source of the input for these components.
We feed the fused embedding $\bm{\Tilde{e}}_m$ along with the duration of the motion sequence $l_m$ into a transformer motion decoder to acquire the reconstructed motion sequence, denoted as $P_{recon}$.
To evaluate the difference between the original and reconstructed motions, we utilize a Mean Squared Error (MSE) loss in training:
\begin{equation}
    \mathcal{L}_{recon} = MSE(P, P_{recon}).
\end{equation}
Our adoption of a customized attention mechanism is motivated by the need to bridge a gap in existing multi-modal alignment methods. These methods often incorporate a reconstruction branch to preserve the integrity of the learned representations in relation to the input data.
The prior methods~\cite{ramesh2022hierarchical, ramesh2021zero} often focus on the information of a single modality, overlooking the rich contextual information available from other modalities. 
For instance, text frequently communicates subtle details that are not immediately apparent in motions or videos, such as intentions, explicit actions, or temporal nuances (\eg, ``quickly'' or ``slowly''). In a similar pattern, videos offer visual context that might not be fully captured by text or motion.
Thus, we draw inspiration from the query, key, and value framework commonly used in natural language processing~\cite{wang2019coke, gao2022calic} to train our model for effective integration of crucial information from each modality. By extracting information from text and video modalities, our model can resolve ambiguities and fill in informational gaps within the motion modality.
Our final training loss can be summarized as:
\begin{equation}
    \mathcal{L} = \mathcal{L}_{align} + {\lambda}_{recon} \cdot \mathcal{L}_{recon},
\end{equation}
where ${\lambda}_{recon}$ is the weight of the reconstruction loss.

\begin{figure}[t]
    \centering
    \includegraphics[width=0.5\textwidth]{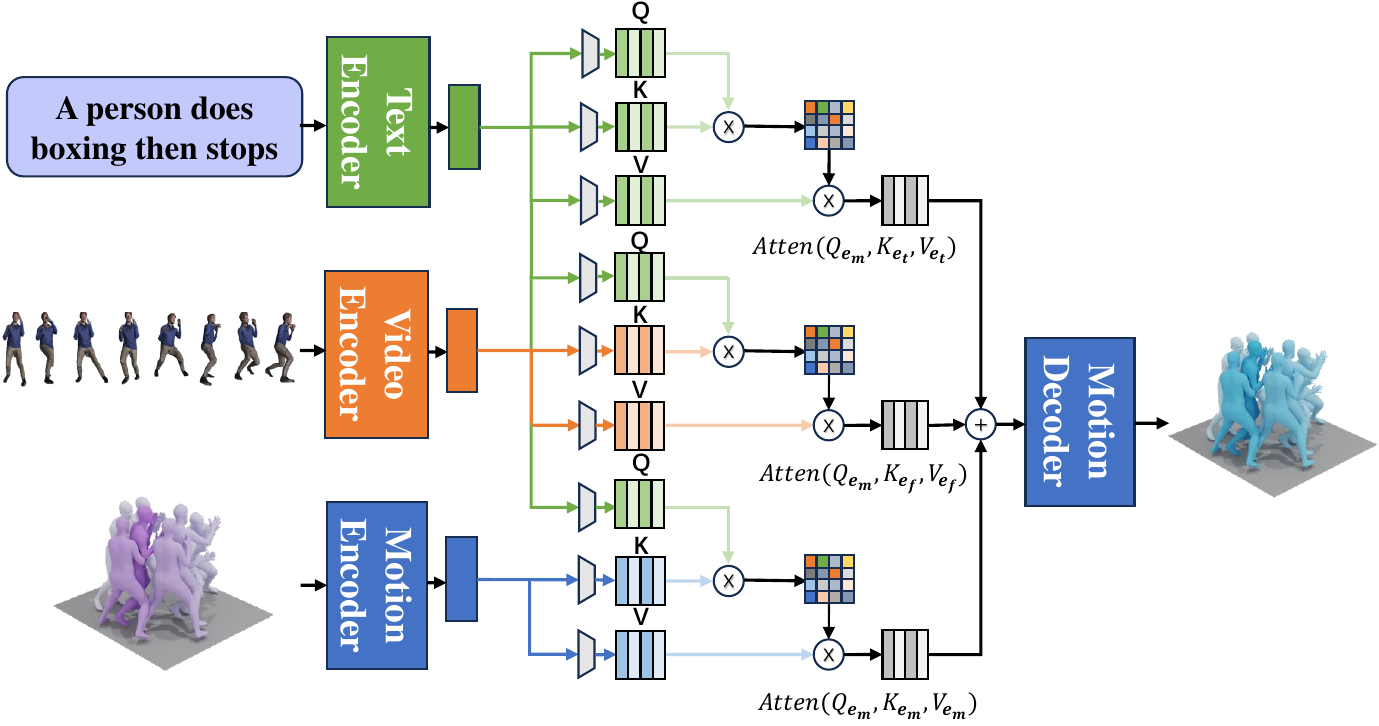}
    \beforefigcaption
    \caption{\textbf{Overview of Features Fusion module.} The embeddings for the text, video, and motion modalities are derived from their respective encoders. Subsequently, the motion embedding acts as a query to retrieve relevant information from the text and video, potentially compensating for any information that may be missing in the motion modality. The output of the attention mechanism is the weighted synthesis of the three modalities, which is then fed to the motion decoder for reconstruction.}
    \afterfigcaption
    \label{fig:2}
\end{figure}


\begin{table*}[t]
    \centering
    \small
    \vspace{-3mm}
    \setlength{\tabcolsep}{1.0pt}
    \begin{tabular}{l|l|*{6}{>{\centering\arraybackslash}m{1.0cm}}|*{6}{>{\centering\arraybackslash}m{1.0cm}}}
        \toprule
        \textbf{Protocol} & \textbf{Methods} & \multicolumn{6}{c|}{\textbf{Text-motion retrieval}} & \multicolumn{6}{c}{\textbf{Motion-text retrieval}} \\
        & & R@1↑ & R@2↑ & R@3↑ & R@5↑ & R@10↑ & MedR↓ & R@1↑ &R@2↑ & R@3↑ & R@5↑ & R@10↑ & MedR↓ \\
        \midrule
        (a) All & TEMOS ~\cite{petrovich2022temos} & 2.12 & 4.09 & 5.87 & 8.26 & 13.52 & 173.00 & 3.86 & 4.54 & 6.94 & 9.38 & 14.00& 183.25 \\
        & Guo et al. ~\cite{Guo_2022_CVPR} & 1.80 & 3.42 & 4.79 & 7.12 & 12.47 & 81.00 &  2.92 & 3.74 & 6.00 & 8.36 & 12.95 & 81.50 \\
        & MotionCLIP ~\cite{tevet2022motionclip} & 2.33 & 5.85 & 8.93 & 12.77 & 18.14 & 103.00 &  5.12 & 6.97 & 8.35 & 12.46 & 19.02 & 91.42 \\
        & TMR~\cite{petrovich23tmr} & 5.68 & 10.59 & 14.04 & 20.34 & 30.94 & 28.00 &  \bf9.95 & 12.44 & 17.95 & 23.56 & 32.69 &28.50 \\
        &Ours(2-modal) & 5.93 & 10.89 & 14.67 & 20.78 & 32.47 & \bf24.00 & 8.08 & 12.30 &   16.85 & 23.39 & 34.64 & 23.50 \\
        & Ours(3-modal) & \bf6.37 & \bf11.84 & \bf15.60 & \bf21.95 & \bf33.67 & \bf24.00 & 9.72 & \bf13.33 & \bf18.73 & \bf25.00 & \bf36.55 & \bf22.50 \\
        \midrule
        (b) All with threshold & TEMOS ~\cite{petrovich2022temos} & 5.21 & 8.22 & 11.14 & 15.09 & 22.12 & 79.00 &  5.48 & 6.19 & 9.00 & 12.01 & 17.10 &129.0 \\
        & Guo et al. ~\cite{Guo_2022_CVPR} & 5.30 & 7.83 & 10.75 & 14.59 & 22.51 & 54.00 & 4.95 & 5.68 & 8.93 & 11.64 & 16.94 &69.50\\
        & MotionCLIP ~\cite{tevet2022motionclip} & 7.22 & 10.58 & 13.48 & 19.07 & 23.65 & 55.00 &  7.10 & 10.21 & 13.57 & 20.04 & 25.44 & 53.87 \\
        & TMR~\cite{petrovich23tmr} & 11.60 & 15.39 & 20.50 & 27.72 & 38.52 & 19.00 & 13.20 & 15.73 & 22.03 & 27.65 & 37.63 & 21.50 \\
        &Ours(2-modal) & 11.32 & 15.01 & 20.91 & 27.93 & 40.97 & 17.00 & 12.95 & \bf17.56 & 22.69 & 29.05 & 39.58 & 19.00 \\
        & Ours(3-modal) & \bf12.94 & \bf17.38 & \bf23.63 & \bf30.13 & \bf42.46 & \bf16.00 & \bf13.89 & 17.11 & \bf23.83 & \bf29.93 & \bf41.09 & \bf17.50\\
        \midrule
        (c) Dissimilar subset & TEMOS ~\cite{petrovich2022temos} & 20.00 & 33.00 & 37.00 & 47.00 & 62.00 & 6.00 & 24.00 & 30.00 & 39,00 & 47.00 & 62.00 & 6.74 \\
        & Guo et al. ~\cite{Guo_2022_CVPR} & 13.00 & 27.00 & 39.00 & 51.00 & 72.00 & 5.00 & 24.00 & 39.00 & 46.00 & 58.00 & 71.00 & 4.50 \\
        & MotionCLIP ~\cite{tevet2022motionclip} & 21.00 & 31.00 & 37.00 & 49.00 & 65.00 & 6.00 &  21.00 & 36.00 & 45.00 & 53.00 & 69.00 & 5.12 \\
        & TMR~\cite{petrovich23tmr} & 34.00 & 56.00 & 61.00 & 68.00 & 76.00 & 2.00 & 47.00 & 55.00 & 65.00 & 71.00 & 78.00 & 2.50 \\
        & Ours(2-modal) & \bf51.00 & 63.00 & 69.00 & 76.00 & 80.00 & 1.50 & 52.00 & 65.00 & 70.00 & 77.00 & 80.00 & \bf1.00 \\
        &Ours(3-modal) & 50.00 & \bf71.00 & \bf79.00 & \bf86.00 & \bf90.00 & \bf1.00 & \bf56.00 & \bf73.00 & \bf81.00 & \bf86.00 & \bf92.00 & \bf1.00 \\
        \midrule
        (d) Small batches ~\cite{Guo_2022_CVPR} & TEMOS ~\cite{petrovich2022temos} & 40.49 & 53.52 & 61.14 & 70.96 & 84.15 & 2.33 & 39.96 & 53.49 & 61.79 & 72.40 & 85.89 & 2.33 \\
        & Guo et al. ~\cite{Guo_2022_CVPR} & 52.48 & 71.05 & 80.65 & 89.86 & 96.58 & 1.39 & 52.00 & 71.21 & 81.11 & 89,87 & 96.78 & 1.38 \\
        & MotionCLIP ~\cite{tevet2022motionclip} & 46.24 & 60.25 & 68.93 & 80.47 & 91.35 & 1.88 &  44.76 & 56.81 & 65.22 & 77.83 & 90.19 & 2.03 \\
        &TMR~\cite{petrovich23tmr} & 67.16 & 81.32 & 86.81 & 91.43 & 95.36 & 1.04 & 67.97 & 81.20 & 86.35 & 91.70 & 95.27 & 1.03  \\
        &Ours(2-modal) & 66.74 & 78.98 & 83.67 & 87.29 & 91.66 & 1.02 & 67.87 & 79.62 & 84.08 & 88.27 & 91.61 & \bf1.01 \\
        &Ours(3-modal) & \bf68.58 & \bf81.04 &\bf85.02 & \bf88.77 & \bf92.58 & \bf1.01 & \bf68.64 & \bf81.06 & \bf85.52 & \bf88.76 & \bf92.82 & \bf1.01 \\
    \bottomrule
    \end{tabular}
    \beforetab
    \caption{\textbf{Text-to-motion Retrieval on HumanML3D.} Presented here are our results on the text-to-motion retrieval tasks conducted on the HumanML3D dataset. The results indicate that our 3-modal version exceeds the performance of Guo et al.~\cite{Guo_2022_CVPR}, TEMOS~\cite{petrovich2022temos}, MotionCLIP~\cite{tevet2022motionclip} and TMR ~\cite{petrovich23tmr}. Moreover, our 3-modal version outperforms our 2-modal version, showing that an extra modality can indeed enhance the alignment between texts and motions. The most notable results are emphasized in \textbf{bold}.}
    \aftertab
    \vspace{-0mm}
    \label{table1}
\end{table*}

\beforesubsection
\subsection{Cross-Modalities Retrieval}
\aftersubsection
\label{sec3.3}
Leveraging the learned joint embedding space across the three modalities, our method achieves various motion-related cross-modal retrieval tasks.

\noindent\textbf{Retrieval Between Text and Motion.} This task is first proposed by TMR~\cite{petrovich23tmr}. The primary objective is to search the most similar motion sequences in response to text queries, or conversely, to find text descriptions that best match given motion sequences. 
The text-to-motion retrieval task offers a unique advantage compared to text-to-motion generation, particularly in scenarios where high authenticity and online efficiency are crucial, such as in animation and game production.

\noindent\textbf{Retrieval Between Video and Motion.} 
This task shares similarities with Human Pose Estimation (HPE) but also addresses several of its key limitations. While HPE has been a groundbreaking technique, it often lacks robustness, which can lead to issues such as jittering or floating feet. These problems can significantly compromise the realism of the motions predicted by HPE.
In contrast, motions retrieved from large-scale motion capture datasets offer a more authentic and realistic experience. These motions are carefully captured from real-world movements, making them highly suitable for applications that demand high fidelity and an accurate representation of human motion.

%% file: sec/4_experiment.tex
\beforesection
\section{Experiment}
\aftersection

\begin{table*}[t]
    \centering
    \small
    \vspace{-1mm}
    \setlength{\tabcolsep}{1.0pt}
    \begin{tabular}{l|l|*{6}{>{\centering\arraybackslash}m{1.0cm}}|*{6}{>{\centering\arraybackslash}m{1.0cm}}}
        \toprule
        \textbf{Protocol} & \textbf{Methods} & \multicolumn{6}{c|}{\textbf{Text-motion retrieval}} & \multicolumn{6}{c}{\textbf{Motion-text retrieval}} \\
        & & R@1↑ & R@2↑ & R@3↑ & R@5↑ & R@10↑ & MedR↓ & R@1↑ &R@2↑ & R@3↑ & R@5↑ & R@10↑ & MedR↓ \\
        \midrule
        (a) All & TEMOS ~\cite{petrovich2022temos} & 7.11 & 13.25 & 17.59 & 24.10 & 35.66 & 24.00 & 11.69 & 15.30 & 20.12 & 26.63 & 36.39 & 26.50 \\
        & Guo et al. ~\cite{Guo_2022_CVPR} & 3.37 & 6.99 & 10.84 & 16.87 & 27.71 & 28.00 & 4.94 & 6.51 & 10.72 & 16.14 & 25.30 & 28.50 \\
        & MotionCLIP ~\cite{tevet2022motionclip} & 4.87 & 9.31 & 14.36 & 20.09 & 31.57 & 26.00 &  6.55 & 11.28 & 17.12 & 25.48 & 34.97 & 23.00 \\
        & TMR~\cite{petrovich23tmr} & 7.23 & 13.98 & 20.36 & 28.31 & 40.12 & 17.00 &  11.20 & 13.86 & 20.12 & 28.07 & 38.55 & 18.00 \\
        &Ours(2-modal) & 8.59 & 15.04 & 21.09 & 32.23 & 46.09 & 13.00 & 11.72 & 17.19 & 23.63 & 32.81 & 48.83 & 12.00 \\
        &Ours(3-modal) & \bf10.16 & {\bf 19.92 } & {\bf24.61} & {\bf34.57} & {\bf49.80} & {\bf11.00} & {\bf15.43} & \bf20.12 & \bf26.95 & \bf34.57 & \bf53.32 & \bf10.00 \\
        \midrule
        (b) All with threshold & TEMOS ~\cite{petrovich2022temos} & 18.55 & 24.34 & 30.84 & 42.29 & 56.37 & 7.00 &  17.71 & 22.41 & 28.80 & 35.42 & 47.11 & 13.25 \\
        & Guo et al. ~\cite{Guo_2022_CVPR} & 13.25 & 22.65 & 29.76 & 39.04 & 49.52 & 11.00 & 10.48 & 13.98 & 20.48 & 27.95 & 38.55 & 17.25\\
        & MotionCLIP ~\cite{tevet2022motionclip} & 13.79 & 23.08 & 31.45 & 42.93 & 53.01 & 9.00 &  13.24 & 22.11 & 29.53 & 38.06 & 50.23 & 10.00 \\
        & TMR~\cite{petrovich23tmr} & 24.58 & 30.24 & 41.93 & 50.48 & 60.36 & 5.00 & 19.64 & 23.73 & 32.53 & 41.20 & 53.01 & 9.50 \\
        &Ours(2-modal) & 24.02 & 30.86 & 42.73 & 54.69 & 70.09 & 5.00 & 21.68 & 27.93 & 34.18 & 42.77 & 57.42 & 8.00\\
        &Ours(3-modal) & \bf30.86 & \bf41.80 & \bf48.63 & \bf59.96 & \bf74.22 & \bf4.00 & \bf25.98 & \bf31.25 & \bf38.28 & \bf45.70 & \bf63.09 & \bf6.50 \\
        \midrule
        (c) Dissimilar subset & TEMOS ~\cite{petrovich2022temos} & 24.00 & 40.00 & 46.00 & 54.00 & 70.00 & 5.00 & 33.00 & 39.00 & 45.00 & 49.00 & 64.00 & 6.50 \\
        & Guo et al. ~\cite{Guo_2022_CVPR}] & 16.00 & 29.00 & 36.00 & 48.00 & 66.00 & 6.00 & 24.00 & 29.00 & 36.00 & 46.00 & 66.00 & 7.00 \\
        & MotionCLIP ~\cite{tevet2022motionclip} & 19.00 & 33.00 & 41.00 & 50.00 & 69.00 & 6.00 &  28.00 & 36.00 & 43.00 & 48.00 & 65.00 & 7.00 \\
        & TMR~\cite{petrovich23tmr} & 26.00 & 46.00 & 60.00 & 70.00 & 83.00 & 3.00 & 34.00 & 45.00 & 60.00 & 69.00 & 82.00 & 3.50 \\
        &Ours(2-modal) & 29.00 & 45.00 & 60.00 & 71.00 & 81.00 & 2.00 & 43.00 & 59.00 & 67.00 & 73.00 & \bf83.00 &\bf2.00 \\
        &Ours(3-modal) & \bf30.00 & \bf49.00 & \bf63.00 & \bf73.00 & \bf84.00 & \bf3.00 & \bf48.00 & \bf60.00 & \bf66.00 & \bf76.00 & 82.00 & \bf2.00 \\
        \midrule
        (d) Small batches ~\cite{Guo_2022_CVPR} & TEMOS ~\cite{petrovich2022temos} & 43.88 & 58.25 & 67.00 & 74.00 & 84.75 & 2.06 & 41.88 & 55.88 & 65.62 & 75.25 & 85.75 & 2.25 \\
        & Guo et al. ~\cite{Guo_2022_CVPR} & 42.25 & 62.62 & 75.12 & 87.50 & 96.12 & 1.88 & 39.75 & 62.75 & 73.62 & 86.88 & 95.88 & 1.95 \\
        & MotionCLIP ~\cite{tevet2022motionclip} & 41.29 & 55.38 & 69.50 & 78.83 & 90.12 & 1.73 &  39.55 & 52.07 & 68.13 & 77.94 & 90.85 & 2.16 \\
        &TMR~\cite{petrovich23tmr} & 49.25 & 69.75 & 78.25 & 87.88 & 95.00 & 1.50 & 50.12 & 67.12 & 76.88 & 88.88 & 94.75 & 1.53  \\
        &Ours(2-modal) & 53.96 & 76.42 & 82.05 & 89.66 & 95.22 & 1.38 & 58.58 & 75.05 & 81.84 & 89.68 & 93.86 &1.23 \\
        &Ours(3-modal) & \bf58.10 & \bf77.80 & \bf86.34 & \bf93.08 & \bf96.47 & \bf1.08 & \bf60.23 & \bf77.52 & \bf86.44 & \bf93.22 & \bf95.87 & \bf1.20 \\
    \bottomrule
    \end{tabular}
    \beforetab
    \caption{\textbf{Text-to-motion Retrieval on KIT-ML.} We conduct further evaluations of both our 2-modal and 3-modal approaches using the KIT-ML dataset. The findings reveal that our 2-modal version significantly surpasses previous methodologies in performance. Moreover, our 3-modal version demonstrates an even greater extent of superiority over other existing methods. The most notable results are emphasized in \textbf{bold}.} 
    \aftertab
    \vspace{-1mm}
    \label{table2}
\end{table*}

We carry out our experiments using the HumanML3D dataset~\cite{Guo_2022_CVPR} and the KIT-ML dataset~\cite{clavet2016motion}. The results demonstrate that our model outperforms previous methods in both the text-to-motion retrieval and video-to-motion retrieval tasks. Our method also exhibits generalization capabilities when applied to real-world, human-centric videos.

\noindent{\bf Datasets.} HumanML3D~\cite{Guo_2022_CVPR} is the largest 3D human motion dataset with text descriptions, featuring 14,616 motions and 44,970 textual descriptions. The texts comprise 5,371 unique words. The motion data, originally from AMASS~\cite{mahmood2019amass} and HumanAct12~\cite{guo2020action2motion}, is downsampled to 20 FPS, cropped to 10 seconds if longer, and oriented to the Z+ direction. Each motion has a minimum of 3 descriptions. The dataset is divided into 80\% training, 5\% validation, and 15\% test sets. KIT-ML~\cite{Plappert2016} is a human motion dataset with 3,911 sequences and 6,278 text annotations, containing a vocabulary of 1,623 unique words. Motions are downsampled to 12.5 FPS. Each sequence has 1 to 4 descriptions, averaging 8 words. The dataset also follows an 80\% training, 5\% validation, and 15\% test split.

\noindent\textbf{RGB Videos for KIT-ML and HumanML3D.} Since the two datasets only contain motion capture data without any RGB videos, we propose to animate avatars and then render RGB videos to augment the video modality for both datasets. For each motion, an avatar is randomly picked from 13 different avatars, then animated and rendered to form its corresponding RGB videos of size $512\times 512$. We include the details of avatars and rendering in the supplementary material.


\noindent\textbf{Implementation Details.} All experiments are conducted using a single NVIDIA A40 GPU with PyTorch~\cite{paszke2017automatic}. Our video encoder is based on ViT-B/32 with 12 layers, complemented by a temporal transformer with 6 layers to consider time series information. The motion encoder comprises a 6-layer transformer encoder, while the text encoder is based on DistilBERT~\cite{DBLP:journals/corr/abs-1910-01108}, supplemented by a temporal transformer to consider the word embedding positions, we finetune DistilBERT during the training process. We follow ~\cite{wu2023revisiting} and initialize DistilBERT and CLIP image encoder trained on the Kinetics-400 dataset. We sample 8 frames from a video sequence, configure the latent dimension $C$ to 512, set $\epsilon$ to 0.8 and assign 0.1 to $\lambda_{recon}$. Training is conducted with a batch size $B$ of 64 over 400 epochs. During the training process, we use AdamW optimizer~\cite{oord2018representation} with a learning rate being 1e-4 and then linearly decaying to 1e-5 after the first 100 epochs. In the process of augmenting data, an image is first resized randomly, from which a crop measuring $256\times 256$ pixels is extracted. Subsequently, this crop is subjected to a variety of transformations, including jittering of colors randomly, conversion to grayscale on a random basis, application of Gaussian Blur, flipping horizontally in a random manner following the implementation of RandAugment~\cite{cubuk2020randaugment}. Our 2-modal version shares the same setting to 3-modal version, with the differences lying in the contrastive learning and modalities fusion between text and motion only.

\noindent{\bf Evaluation Metrics.} Our evaluation of retrieval performance utilizes standard metrics, including recall at various ranks (\eg, R@1, R@2) for both text-motion and video-motion tasks. A higher R-precision value indicates a more accurate retrieval. Additionally, we assess the median rank (MedR) of our results. MedR represents the median ranking position of the ground-truth result, with lower values indicating more precise retrievals. Following TMR~\cite{petrovich23tmr}, the four used evaluation protocols are outlined below: (i) {\bf All} uses the entire test dataset as the retrieval database. However, the precision may be compromised as texts categorized into negative pairs could still convey similar meanings. (ii) {\bf All with threshold} addresses the problem mentioned above, we set the threshold to 0.8, same to the negative filtering threshold. If the similarity between the retrieved motion and the ground-truth motion exceeds this threshold, the result is deemed accurate. (iii) {\bf Dissimilar subset} retrieves motion from a refined subset. The database comprises 100 sampled pairs, with each pair being distinctly dissimilar. Consequently, it's relatively easier to retrieve the correct motion with this protocol compared to the prior ones. (iv) {\bf Small batches} involves randomly selecting batches of 32 motion-text pairs and assessing the average performance. 
\label{sec4.1}

\begin{figure*}
    \centering
    \vspace{-2mm}
    \includegraphics[width=\textwidth]{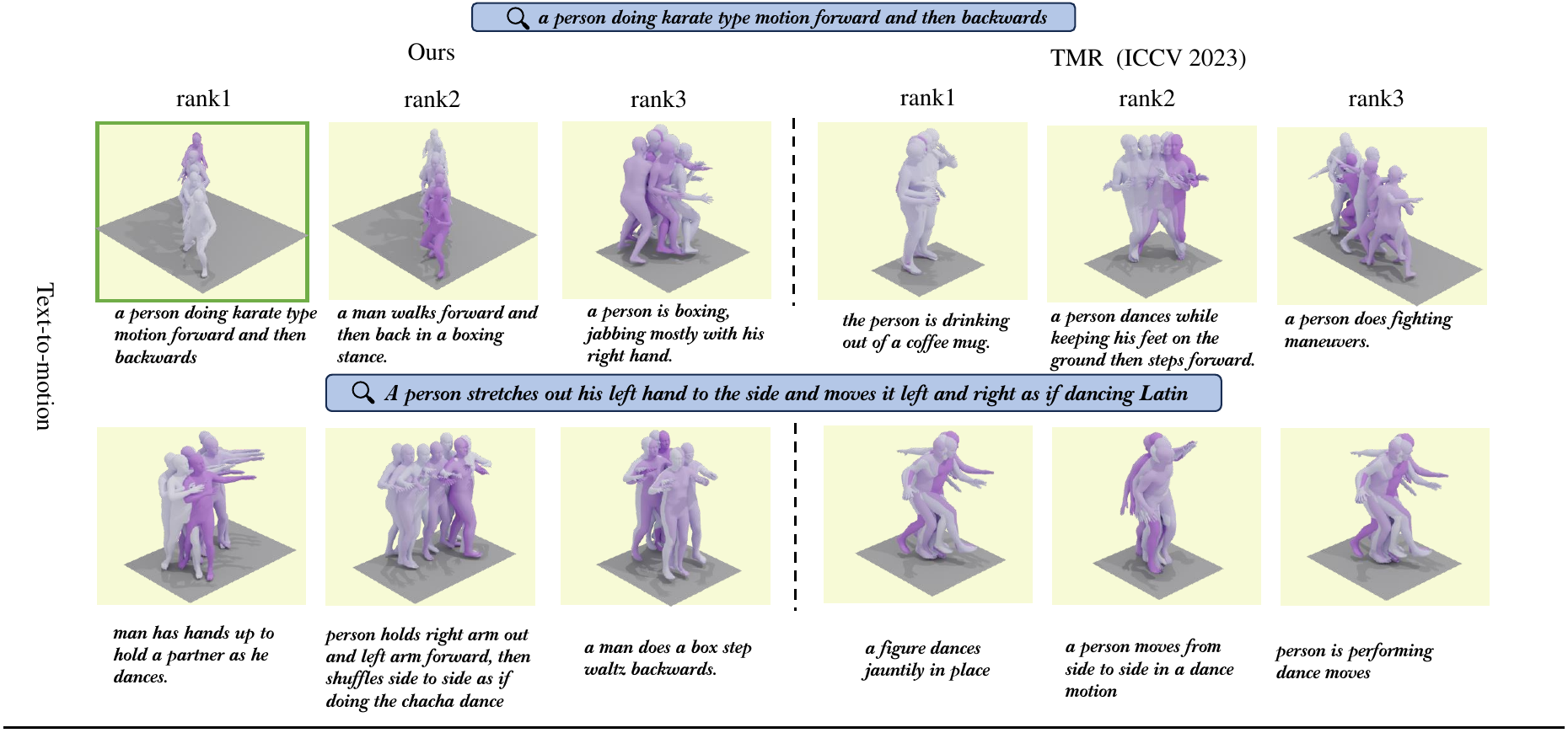}
    \includegraphics[width=\textwidth]{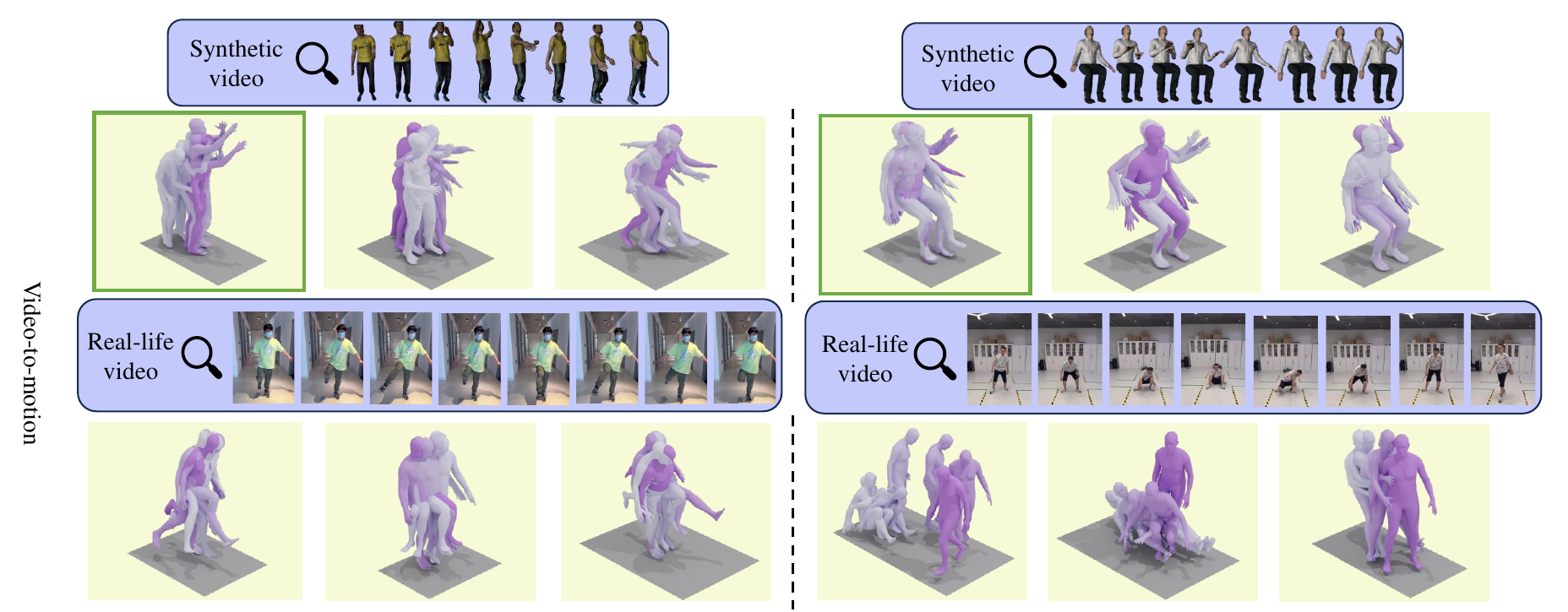}
    \beforefigcaption
    \caption{\textbf{Qualitative Comparison on the HumanML3D Dataset.} Our method successfully performs text-to-motion and video-to-motion retrieval tasks. For text-to-motion retrieval, we compare our results with TMR~\cite{petrovich23tmr}. In the first row, using a random text from the test set, our method accurately retrieves the correct motion at rank 1, with similar motions such as ``boxing'' at ranks 2 and 3, resembling '`karate type motion''. In contrast, TMR struggles, with only its rank 3 motion matching the ground truth. In the second row, when testing with a non-test set text involving ``dance'', our model retrieves motions suggesting a ``Latin dance'', more accurate than TMR's less precise dance motions. For video-to-motion retrieval, in the third row, our model excels with test set videos, correctly retrieving ground-truth motions at rank 1 and similar motions at ranks 2 and 3. Furthermore, in the last row, when applied to real-life human-centric videos, our model shows strong generalization, retrieving motions closely matching the video content, such as ``leg swinging'' and ``standing up and walking''.}
    \afterfigcaption
    \label{fig:3}
\end{figure*}

\begin{table*}[t]
    \centering
    \small
    \vspace{-3mm}
    \setlength{\tabcolsep}{1.0pt}
    \begin{tabular}{l|l|*{6}{>{\centering\arraybackslash}m{1.0cm}}|*{6}{>{\centering\arraybackslash}m{1.0cm}}}
        \toprule
        \textbf{Protocol} & \textbf{Methods} & \multicolumn{6}{c|}{\textbf{Video-motion retrieval}} & \multicolumn{6}{c}{\textbf{Motion-video retrieval}} \\
        & & R@1↑ & R@2↑ & R@3↑ & R@5↑ & R@10↑ & MedR↓ & R@1↑ &R@2↑ & R@3↑ & R@5↑ & R@10↑ & MedR↓ \\
        \midrule
        (a) All & MotionCLIP ~\cite{tevet2022motionclip} & 4.96 & 9.40 & 12.48 & 17.46 & 26.46 & 43.00 & 3.52 & 7.28 & 9.74 & 13.82 & 22.34 & 58.00\\
        &MotionSet~\cite{ren2020video} & 8.26 & 18.69 & 35.83 & 48.91 & 56.32 & 16.00 & 10.17 & 21.33 & 37.39 & 51.25 & 58.97 & 13.00\\
        &Ours(3-modal) & \bf21.25 & \bf41.38 & \bf50.51 & \bf63.40 & \bf77.50 & \bf3.00 & \bf25.76 & \bf45.36 & \bf54.71 & \bf66.87 & \bf79.73 & \bf3.00\\
        \midrule
        (b) All with threshold & MotionCLIP ~\cite{tevet2022motionclip} & 7.10 & 12.62 & 16.67 & 22.56 & 32.23 & 29.00 & 5.71 & 9.20 & 12.35 & 17.41 & 28.0 &38.00\\
        &MotionSet~\cite{ren2020video} & 15.63 & 31.17 & 40.22 & 52.93 & 65.47 & 8.00 & 17.68 & 35.82 & 43.11 & 54.25 & 66.99 & 7.00\\
        &Ours(3-modal) & \bf30.42 & \bf46.97 & \bf56.82 & \bf68.77 & \bf81.41 & \bf3.00 & \bf35.43 & \bf52.73 & \bf61.93 & \bf72.95 & \bf84.54 &\bf2.00\\
        \midrule
        (C)Dissimilar subset & MotionCLIP ~\cite{tevet2022motionclip} & 42.00 & 61.00 & 72.00 & 80.00 & 88.00 & 2.00 & 43.00 & 65.00 & 69.00 & 76.00 & 86.00 &2.00\\
        &MotionSet~\cite{ren2020video} & 58.00 & 69.00 & 81.00 & 89.00 & 97.00 & 2.00 & 59.00 & 73.00 & 86.00 & 90.00 & 98.00 & 2.00\\
        &Ours(3-modal) & \bf74.00 & \bf93.00 & \bf98.00 & \bf100.00 & \bf100.00 & \bf1.00 & \bf79.00 & \bf99.00 & \bf100.00 & \bf100.00 & \bf100.00 &\bf1.00\\
    \bottomrule
    \end{tabular}
    \beforetab
    \caption{\textbf{Video-to-motion Retrieval on HumanML3D.} We assess the video-to-motion retrieval task using the HumanML3D datasets. Our approach surpasses MotionCLIP~\cite{tevet2022motionclip} and MotionSet~\cite{ren2020video} across all the evaluation protocols in the table. The most notable results are emphasized in \textbf{bold}.}
    \aftertab
    \label{table3}
\end{table*}

\begin{table*}[t]
    \centering
    \small
    \setlength{\tabcolsep}{1.0pt}
    \begin{tabular}{l|l|*{6}{>{\centering\arraybackslash}m{1.0cm}}|*{6}{>{\centering\arraybackslash}m{1.0cm}}}
        \toprule
        \textbf{Protocol} & \textbf{Methods} & \multicolumn{6}{c|}{\textbf{Video-motion retrieval}} & \multicolumn{6}{c}{\textbf{Motion-video retrieval}} \\
        & & R@1↑ & R@2↑ & R@3↑ & R@5↑ & R@10↑ & MedR↓ & R@1↑ &R@2↑ & R@3↑ & R@5↑ & R@10↑ & MedR↓ \\
        \midrule
        (a) All & MotionCLIP ~\cite{tevet2022motionclip} & 3.71 & 6.45 & 8.79 & 11.52 & 17.97 & 52.50 & 4.49 & 7.23 & 9.96 & 12.11 & 21.68 & 51.00\\
        &MotionSet~\cite{ren2020video} & 9.15 & 23.96 & 35.27 & 46.18 & 61.31 & 13.00 & 11.08 & 24.27 & 35.03 & 45.29 & 60.53 & 13.00\\
        &Ours(3-modal) & \bf18.75 & \bf35.16 & \bf42.58 & \bf55.86 & \bf73.24 & \bf4.00 & \bf23.05 & \bf35.55 & \bf44.14 & \bf56.25 & \bf72.07 & \bf4.00\\
        \midrule
        (b) All with threshold & MotionCLIP ~\cite{tevet2022motionclip} & 8.98 & 12.11 & 14.84 & 19.53 & 27.73 & 27.00 & 9.18 & 12.50 & 17.19 & 22.27 & 33.98 & 21.00\\
        &MotionSet~\cite{ren2020video} & 20.17 & 35.21 & 43.38 & 55.39 & 69.42 & 6.00 & 24.57 & 37.66 & 48.29 & 59.72 & 70.71& 5.00\\
        &Ours(3-modal) & \bf36.91 & \bf49.80 & \bf60.94 & \bf70.70 & \bf84.57 & \bf3.00 & \bf41.41 & \bf52.73 & \bf63.28 & \bf75.98 & \bf86.13 & \bf2.00\\
        \midrule
        (C)Dissimilar subset & MotionCLIP ~\cite{tevet2022motionclip} & 15.00 & 23.00 & 31.00 & 40.00 & 59.00 & 8.00 & 13.00 & 21.00 & 25.00 & 41.00 & 63.00 & 6.00\\
        &MotionSet~\cite{ren2020video} & 40.00 & 58.00 & 73.00 & 85.00 & 89.00 & 2.00 & 40.00 & 60.00 & 73.00 & 82.00 & 87.00 & 2.00\\
        &Ours(3-modal) & \bf66.00 & \bf84.00 & \bf90.00 & \bf96.00 & \bf99.00 & \bf1.00 & \bf66.00 & \bf87.00 & \bf93.00 & \bf98.00 & \bf100.00 & \bf1.00\\
    \bottomrule
    \end{tabular}
    \beforetab
    \caption{\textbf{Video-to-motion Retrieval on KIT-ML.} We assess the video-to-motion retrieval task using the KIT-ML datasets. Our findings align with those obtained from the HumanML3D, indicating that our framework significantly outperforms the performance of MotionCLIP~\cite{tevet2022motionclip} and MotionSet~\cite{ren2020video} by a considerable margin.}
    \aftertab
    \label{table4}
\end{table*}

\beforesubsection
\subsection{Comparison to Prior Works}
\aftersubsection
In this section, we compare our framework to the prior works on text-to-motion retrieval and video-to-motion retrieval tasks.
For the tasks associated with text-to-motion retrieval, we introduce two distinct versions of our framework: the 2-modality version which is only trained on text and motion modalities, and the 3-modality version which is trained on all the 3 modalities. We compare our proposed methodologies against previous methods, including Guo et al.~\cite{Guo_2022_CVPR}, TEMOS~\cite{petrovich2022temos}, MotionCLIP~\cite{tevet2022motionclip} and TMR~\cite{petrovich23tmr}. The comparative results of these evaluations are tabulated in Table \ref{table1} and Table \ref{table2}. Upon assessing performance on the HumanML3D dataset, it shows that our 3-modality framework surpasses the current state-of-the-art TMR across all evaluation protocols for both text-to-motion and motion-to-text retrieval tasks. Our three-modality framework also outperforms the two-modality one, demonstrating that the additional video modality can indeed enhance the alignment between text and motion modalities. On the KIT-ML dataset, we observe a even larger margin compared to HumanML3D dataset.
In our new proposed video-to-motion retrieval tasks, we benchmark our work against MotionCLIP~\cite{tevet2022motionclip} and MotionSet~\cite{ren2020video}. MotionCLIP stands out as the sole work we have identified that leverages an extra modality to boost outcomes in the realm of human motion. However, MotionCLIP uses one randomly selected RGB frame as the input, rather than a sequence of video frames, ignoring temporal features of motion sequences in the video modality. MotionSet converts the original video or MoCap clips to binary silhouette sets and extracting the motion features by a MotionSet network. Given that MotionCLIP is initially trained on the BABEL dataset~\cite{punnakkal2021babel} and MotionSet is initially trained on a self-made dataset, we retrain both the models on the HumanML3D and KIT-ML datasets for equitable comparison. Furthermore, considering MotionCLIP's incorporation of images as an auxiliary modality, we conduct a fair comparison by sampling eight images and applying mean pooling to the model's outputs. Our evaluations, as detailed in Table~\ref{table3} and Table~\ref{table4}, reveal that our 3-modality framework significantly surpasses the performance of MotionCLIP and MotionSet on both the HumanML3D and KIT-ML datasets.
\label{sec4.2}

\beforesubsection
\subsection{Qualitative Results}
\aftersubsection

Figure~\ref{fig:3} presents the visual results for the test subset within the HumanML3D dataset \cite{Guo_2022_CVPR}. In the text-to-motion retrieval task, our framework is compared against the TMR method \cite{petrovich23tmr}. By selecting a textual description randomly from the database, our model demonstrates precision in retrieving the exact ground-truth motion in the rank1 position. Subsequent motions, ranking second and third, while not perfect matches, exhibit similar key motion elements, including boxing and pacing activities. Moreover, when a text prompt describing a dance motion contained in the dataset is inputted, our model consistently identifies the top three corresponding motions, all indicative of Latin dance styles. In the video-to-motion retrieval task, we feed two rendered videos into the model. The first video features a character dribbling and shooting a basketball, while the second shows a character sitting and performing a ``No'' gesture. Our model retrieves the precise ground-truth motion in rank1 position, showing its effectiveness in the video-to-motion retrieval task. In an additional assessment, when two real-life videos are fed into the model, the top three retrieved motions accurately reflect the actions depicted in the videos, demonstrating our model's generalization ability to real-life videos. 
\label{sec4.3}

\beforesubsection
\subsection{User study: Video-to-Motion Retrieval}
\aftersubsection
We create a small dataset comprising 100 real-life videos. The actions in these videos are performed in imitation of the motions found in the test sets of HumanML3D~\cite{Guo_2022_CVPR} and KIT-ML~\cite{Plappert2016}. Specifically, we select 80 motions from HumanML3D and 20 from KIT-ML for imitation. To enhance  diversity, we engage 5 performers, each with a distinctly different body type. We instruct users to view a real-life video and then choose the most similar video from four provided videos. Among these four videos, one is the top-ranked motion retrieved by our framework, while the other three are randomly selected from the corresponding dataset. Additionally, we offer a ``None of the above'' option for users in case they believe the real-life video does not closely match any of the provided options. The results show that our framework successfully retrieves 3D motions that are subjectively similar for 68.5\% of the videos, demonstrating its effective generalization capabilities to real-life video scenarios. 
Our user study web page can be found in \url{https://lavimo2023.github.io/User-Study-LAVIMO/}
\label{sec4.4}

\beforesubsection
\subsection{Limitations}
\aftersubsection
Although LAVIMO demonstrates enhanced efficacy in both text-to-motion and video-to-motion retrieval tasks, there are limitations that future research should consider addressing. Firstly, the video modality in our approach is derived from animating and rendering avatars to match specific motions, which inherently deviates from authentic human-centric videos. While our framework exhibits a degree of generalization to real-life video content, the substitution of rendered footage with actual human-centric videos in our dataset may yield improvements. Secondly, despite outperforming existing benchmarks, such as the model presented by TMR~\cite{petrovich23tmr}, our model does not yet match the precision achieved in video and image retrieval tasks. This discrepancy primarily stems from the limited availability of comprehensive motion datasets, which constrains the broader application potential of the motions retrieved, including their integration into the motion generation pipeline.
\label{sec4.5}

%% file: sec/5_conclution_discussion.tex
\beforesection
\section{Conclusion and Discussion}
\aftersection
In this paper, we introduce LAVIMO, a Unified Language-Video-Motion Alignment framework. By employing contrastive learning, we effectively integrate three distinct modalities into a cohesive embedding space. Our approach surpasses previous methodologies in text-to-motion retrieval tasks, as evidenced by our results on the HumanML3D and KIT-ML datasets. Furthermore, we propose a novel video-to-motion retrieval task, which may serve as an adjunct to existing video pose estimation techniques. Additionally, we demonstrate the capacity of our framework to generalize effectively to real-life human-centric video content.